\documentclass[journal]{IEEEtran}
\usepackage{multirow}
\usepackage{floatrow}
\newfloatcommand{capbtabbox}{table}[][\FBwidth]
\usepackage{amsmath}
\usepackage{amssymb}
\usepackage{caption}
\usepackage{hyperref}

\ifCLASSINFOpdf
\else
\fi
\usepackage{graphicx}

\begin{document}
%
\title{DLGAN: Disentangling Label-Specific\\ Fine-Grained Features for Image Manipulation} 
%
%
%

\author{Guanqi~Zhan*, Yihao~Zhao*, Bingchan~Zhao, Haoqi~Yuan, 
Baoquan~Chen and Hao~Dong\\
CFCS, Computer Science Dept., Peking University\\
\thanks{*equal contribution}
}


\maketitle

\begin{abstract}
\textbf{
Recent studies have shown how disentangling images into content and feature spaces can provide controllable image translation/ manipulation. 
    In this paper, we propose a framework to enable utilizing \emph{discrete} multi-labels to control which features to be disentangled, \emph{i.e.,} disentangling label-specific fine-grained features for image manipulation (dubbed DLGAN). 
    By mapping the \emph{discrete} label-specific attribute features into a \emph{continuous} prior distribution, we leverage the advantages of both discrete labels and reference images to achieve 
    image manipulation in a hybrid fashion.
    For example, given a face image dataset (e.g., CelebA) with multiple discrete fine-grained labels, we can learn to \emph{smoothly interpolate} a face image between black hair and blond hair through reference images while \emph{immediately controlling} the gender and age through discrete input labels.
    To the best of our knowledge, this is the first work that realizes such a hybrid manipulation within a single model.
    More importantly, it is the first work to achieve 
    image interpolation between two different domains without requiring continuous labels as the supervision.
    Qualitative and quantitative experiments demonstrate the effectiveness of the proposed method.
    }
\end{abstract}

\textbf{
}

%
\IEEEpeerreviewmaketitle

\section{Introduction}
Image-to-image translation focuses 
on synthesizing images conditioned on the semantic visual information of other images. 
It has a wide range of applications, such as image super-resolution~\cite{Ledig2017photo}, image colorization~\cite{zhang2016colorful}, artwork synthesis~\cite{Translation2019}, and photo-realistic image synthesis~\cite{wang2018high}.
Pix2pix~\cite{Isola2017} achieves image-to-image translation in a supervised setting, and CycleGAN~\cite{zhu2017cyclegan} achieves the translation without requiring paired images for supervision.

To provide a more controllable synthesis, there are many works that manipulate/translate the input image by leveraging additional information, such as discrete labels~\cite{Choi2018starGAN}, text descriptions~\cite{dong2017semantic}, bash-based interfaces~\cite{Bau2019} and style images~\cite{Xun2017StyleTrans}.
They enable users to manipulate the image in a different way.
On the other hand, other works, such as DRIT~\cite{Lee2019DRIT} and MUNIT~\cite{Huang2018MUINT} disentangle the latent space of the image into a domain-invariant content space and a domain-specific attribute space. 
They can translate an image from one domain to another conditioned on the style of the reference image.

\begin{figure}[h]
    \begin{center}
        \includegraphics[scale=0.5, trim={4cm, 5.1cm, 1cm, 3cm}, clip]{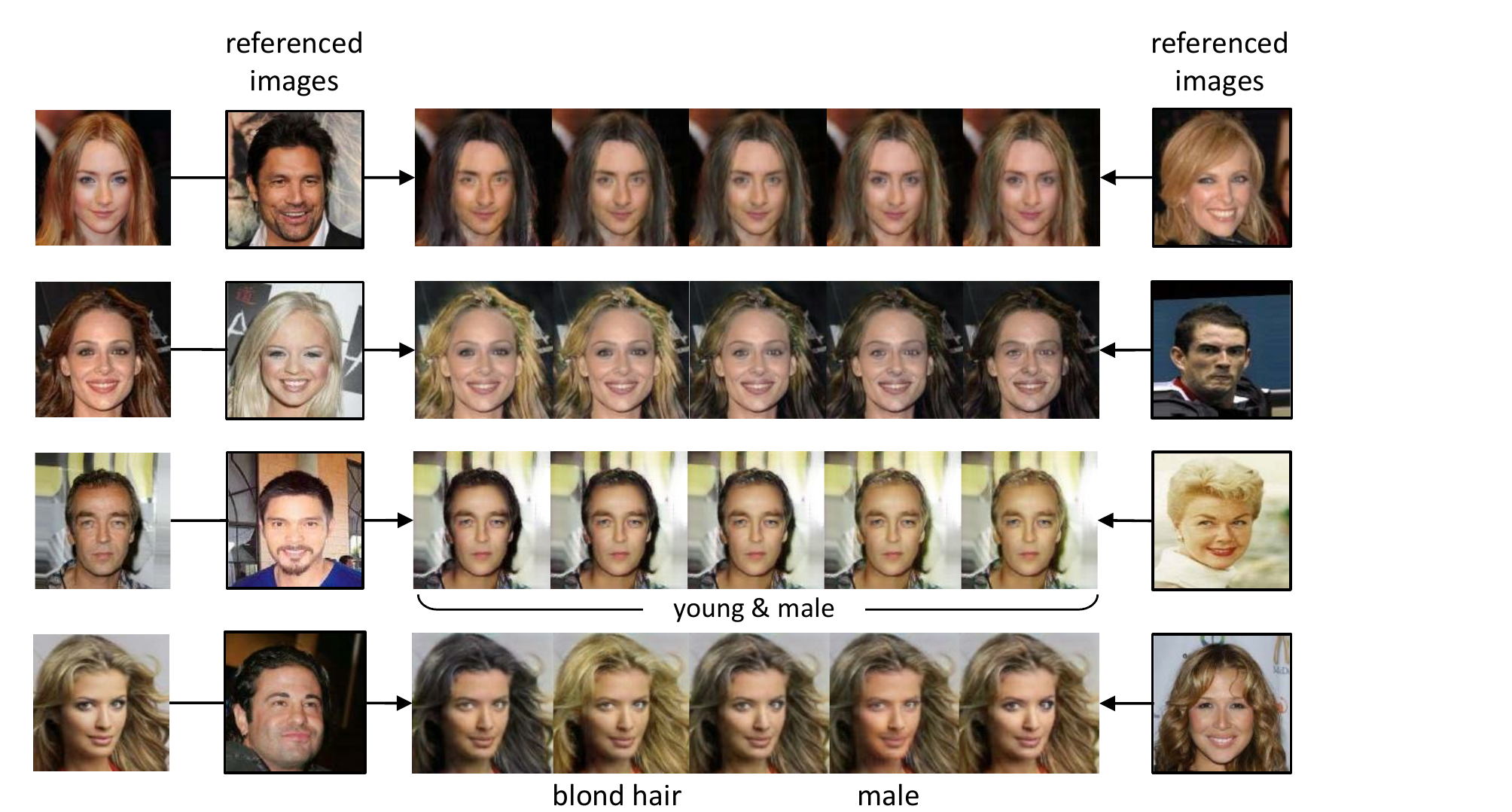}
    \end{center}
    \caption{
    \textbf{DLGAN achieves image interpolation between two domains without requiring continuous labels.} Reference images control the \emph{smooth manipulation} while discrete labels control the \emph{immediate manipulation}:
    The first two rows show examples of smoothly manipulating the hair color and gender, while maintaining the other content information (\emph{e.g.,} the background).
    The last two rows show the labels taking precedence over the reference images.
    }
    \label{fig:homepage}
\end{figure}

Different from those works, this paper focuses on leveraging the advantages of both fine-grained discrete labels and reference images to achieve a more challenging task: interpolation between two domains without continuous labels as supervision. 
By first using the \emph{discrete} fine-grained labels to control which features to be disentangled, 
and then mapping the \emph{discrete} fine-grained features into a \emph{continuous prior distribution}, as shown in Fig.~\ref{fig:homepage}, we enable \emph{smooth linear interpolation} between the reference images from different domains. Meanwhile, additional labels can be used as a constraint to enable \emph{immediate manipulation}, providing a hybrid and fine-grained way to manipulate images. We call the proposed method as Disentangle Label GAN (DLGAN).

To sum up, the main contributions and highlight of our work are:
\begin{itemize}
    \item 
    We utilize discrete multi-labels to control which features to be disentangled, and enable interpolation between two domains without using continuous labels.
    \item 
    We propose an end-to-end method to support image manipulation conditioned on both images and labels, enabling both smooth and immediate changes simultaneously.
    \item 
    Experiments on CelebA~\cite{liu2015faceattributes} and DukeMTMC~\cite{Ristani2016performance} datasets demonstrate the effectiveness of the proposed method compared with baseline methods.
\end{itemize}


    

\section{Related Work}

\begin{table*}[t]
\centering
\begin{tabular}{|l|l|l|l|l|l|l|l|l|}
\hline
                                                                                                                                & \rotatebox{60}{\scriptsize CycleGAN} & \rotatebox{60}{\scriptsize MUNIT} & \rotatebox{60}{\scriptsize DRIT} & \rotatebox{60}{\scriptsize EG-UNIT} &  \rotatebox{60}{\scriptsize StarGAN} & \rotatebox{60}{\scriptsize STGAN} & \rotatebox{60}{\scriptsize GANimation} & \rotatebox{60}{\textbf{ours}} \\ \hline
Multi-modal translation  &&$\surd$&$\surd$ &$\surd$ &&&&$\surd$  \\ \hline
Multi-label translation &&&&&$\surd$ &$\surd $&$\surd$&$\surd$ \\ \hline
Reference image  & &$\surd$ &$\surd$ &$\surd$ &&&&$\surd$   \\ \hline
Disentanglement of domain-specific features & &$\surd$&$\surd$ &$\surd$ && & &$\surd$  \\ \hline
Disentanglement of label-specific features &&&&&&& &$\surd$   \\ \hline
\begin{tabular}[c]{@{}l@{}}Interpolation between two different domains \\ using discrete labels as the supervision\end{tabular} &&&&&&&&$\surd$      \\ \hline
\end{tabular}
 \caption{\textbf{Comparison against existing methods.} CycleGAN~\cite{zhu2017cyclegan}, MUNIT~\cite{Huang2018MUINT}, DRIT~\cite{Lee2018DRIT}, EG-UNIT~\cite{ma2019EG-UNIT},  StarGAN~\cite{Choi2018starGAN}, STGAN~\cite{Liu2019STGAN}, GANimation~\cite{Puma2018GANimation}, and DLGAN (Ours)
 }
    \label{table:novelty}
\end{table*}

\subsection{Generative Adversarial Networks (GAN)} 
Generative Adversarial Networks learn to map one distribution to another in a feed-forward pass~\cite{goodfellow2014gan}~\cite{Karras2019stylegan}. 
Conditional GAN is an important species of GAN that can generate images conditioned on input information.
Prior works use conditioned GANs to synthesis images using different kinds of information, such as discrete labels~\cite{denton2015deep,odena2016acgan,mirza2014conditional}, text captions~\cite{reed2016generative,zhang2017stackgan}, and images~\cite{Isola2017}.

\subsection{Image-to-Image Translation} 
Image-to-image translation usually uses conditional GANs that synthesize images conditioned on the input images. 
Pix2pix~\cite{Isola2017} and BicycleGAN~\cite{zhu2017toward} achieve great success 
when paired images are accessible for training.
There have also been works dealing with unpaired images, such as CycleGAN~\cite{zhu2017cyclegan}, DiscoGAN~\cite{kim2017discogan}, and UNIT~\cite{liu2017unsupervised}.
Image manipulation can be further controlled by additional auxiliary information, such as text caption~\cite{dong2017semantic}, discrete label~\cite{Choi2018starGAN,Liu2019STGAN}, style image~\cite{Xun2017StyleTrans}, bash-based interface~\cite{Bau2019}, Action Unit annotations~\cite{Puma2018GANimation}, and sparse annotation~\cite{reed2016learning}. 
For multi-domain image-to-image translation, StarGAN~\cite{Choi2018starGAN} enables flexible translation for an image to any desired target domain with a label loss applied to the discriminator during the training process. GANimation~\cite{Puma2018GANimation} utilizes fine-grained annotations in a continuous manifold to achieve translation among different domains. 
In STGAN~\cite{Liu2019STGAN}, selective transfer units are incorporated with encoder-decoder to adaptively select and modify encoder feature for enhanced attribute editing, achieving multi-domain image-to-image translation.
In contrast to those works, we use GAN conditioned on both discrete labels and reference images for controllable image manipulation, enabling both smooth and immediate changes simultaneously.

\subsection{Feature Disentanglement.}
Disentangling an image into different constituents requires modeling the factors of data variations. 
There have been early works disentangling images into class-related and class-independent components~\cite{Kingma2014semi,Cheung2015Discover,Mathieu2016Disentangling,Makhzani2016Adversarial}.
As for unsupervised learning, InfoGAN~\cite{chen2016infogan} maximizes the mutual information between some latent variables and observation to achieve disentanglement.
For unpaired image-to-image translation (\emph{i.e.,} only one binary label), DRIT~\cite{Lee2019DRIT}, MUNIT~\cite{Huang2018MUINT}, and EG-UNIT~\cite{ma2019EG-UNIT} decompose the image representation into a content feature space that is domain-invariant and a style space that captures domain-specific properties to achieve the translation of an image to another domain.
Our work differs from all these works in disentangling the label-specific fine-grained features 
and enabling the interpolation from one domain to others rather than the interpolation within a single domain.

Table~\ref{table:novelty} compares DLGAN with the most related works. Using both discrete multi-labels and reference images, our method achieves disentanglement of label-specific features, enabling interpolation between two different domains without continuous labels as the supervision.

\section{Method}

\begin{figure*}[htb]
    \begin{center}
        \includegraphics[scale=0.7, trim={2cm, 6cm, 3cm, 5cm}, clip]{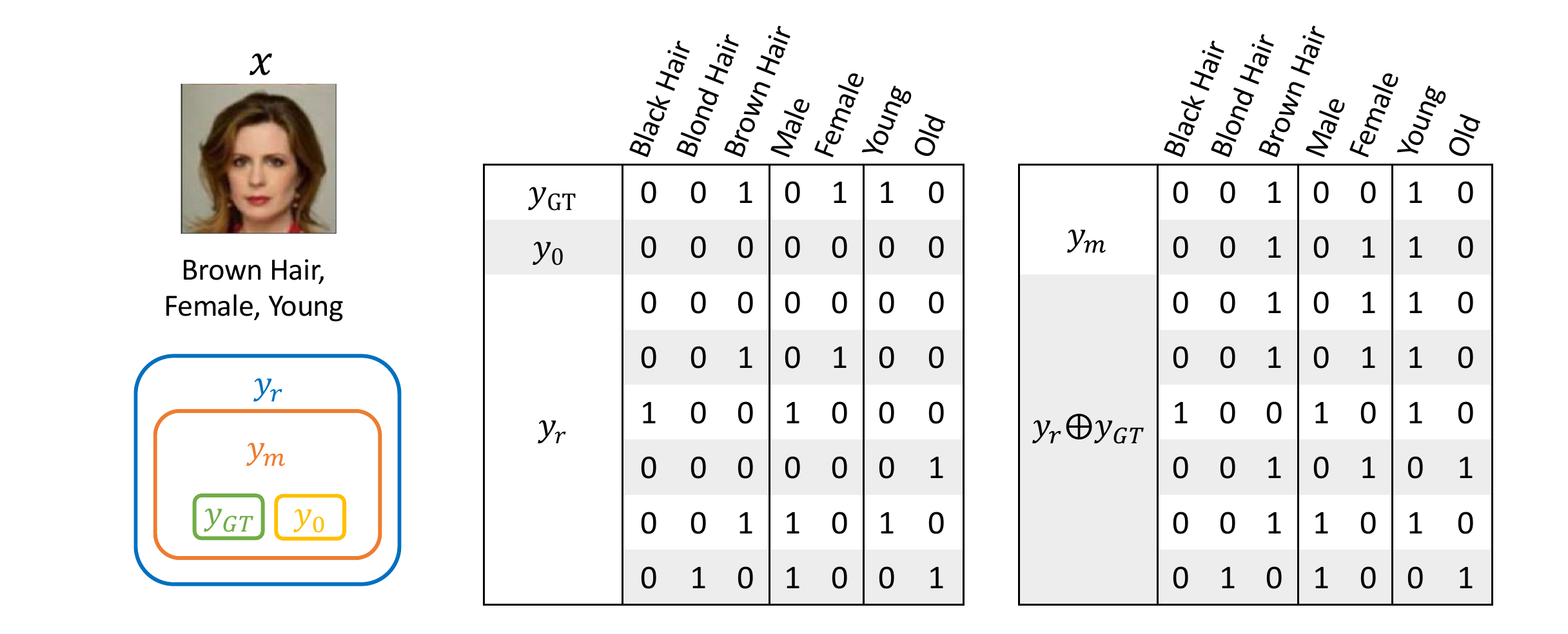}
        \caption{\textbf{Example labels of a CelebA sample with the features of hair color, gender, and age.} 
        The relationships of different types of labels are shown in the Venn diagram on the bottom-left. 
        }
        \label{fig:explainlabel}
    \end{center}
\end{figure*}


Let $(\mathcal{X}, \mathcal{Y}) \subset (\mathbb{R}^{H\times W\times 3}, \mathbb{Y})$ be the space of a certain domain of images and labels, $x\in \mathcal{X}$ be an image sample and $y\in \mathcal{Y}$ be the corresponding discrete multi-labels that describe the fine-grained attributes of the image.
The feature space of the image can be divided into two parts: 1) the label-specific attribute space $A$, which consists of the fine-grained features that the label $y$ mentions, and 2) the content space $C$, which contains content features that the label $y$ does not mention, such as the background and object shape.
For example, in CelebA~\cite{liu2015faceattributes}, if $\mathcal{Y}=\{hair color, gender, age\}$, then the label-specific attribute space $A$ consists of the attribute features of hair color, gender and age while the content space $C$ contains other not mentioned features, including the background as well as the shape of people's face.
%
%
Our goal is to use the discrete multi-labels to control which attribute features to be disentangled and map the attribute features into a continuous prior distribution. In doing so, we enable the interpolation from one domain to the others. 

There are four networks in our model: an attribute encoder $E_a$, a content encoder $E_c$, a generator $G$, and a discriminator $D$. Their target can be viewed as four mappings in different spaces. The attribute encoder $E_a$ maps images into the label-specific attribute space ($E_a: \mathcal{X} \rightarrow A$) and the content encoder $E_c$ maps images into the content space ($E_c: \mathcal{X} \rightarrow C$). The generator $G$ generates images conditioned on the content features, attribute features, and the input labels ($G: \{C,A,\mathcal{Y} \} \rightarrow \mathcal{X}$). The discriminator $D$ aims to classify the input images into correct labels as well as to judge whether the images are realistic ($D: \mathcal{X} \rightarrow \{ \{0,1\},\mathcal{Y}\}$, where the binary variable represents whether the image is real or fake). 
%

\subsection{Fine-Grained Label Representation}
\label{ch:label}


To achieve image manipulation and feature disentanglement,
we define five different ways to represent the discrete fine-grained labels. 
An example of a certain image $x$ is shown in Fig.~\ref{fig:explainlabel}, where 1) $y_{GT}$ denotes the ground-truth label where all feature information is provided,
2) $y_0$ denotes the empty label, which is a zero vector without feature information, 
3) $y_m$ denotes the matching label, where some of the ground truth features or even all of them are missing, and the remaining features must match the image,
4) $y_r$ denotes the random label, which includes all possible combinations of the features and some or all features can be missing,
and 5) $y_r\oplus y_{GT}$ denotes another type of random label where the missing features of $y_r$ are filled with the ground truth features. 

Concretely, take CelebA~\cite{liu2015faceattributes} as an example, if $\mathcal{Y}=\{hair color, gender, age\}$, each label $y$ can be represented by seven bits, where the first three bits represent the information of hair colors: ``100'', ``010'', and ``001'' indicate black, blond, and brown hair, respectively, and ``000'' means missing the information of hair color.
The fourth and fifth bits represent the gender information, where ``10'' and ``01'' represent male and female, respectively, and ``00'' means missing the information of gender.
Similarly, the last two bits represent the age information, where ``10'', ``01'', and ``00'' represent young, old, and unknown, respectively.
More examples and details of the label representation can be found in the appendix.
The missing-feature property of $y_m$ and $y_r$ is important for our feature disentanglement. Details will be discussed in Section~\ref{ch:disentanglement}.



\subsection{Adversarial Image Manipulation}
\label{ch:adversarial}
\begin{figure*}[t]
    \begin{center}
        \includegraphics[scale=0.6, trim={3cm, 6.5cm, 2.8cm, 7cm}, clip]{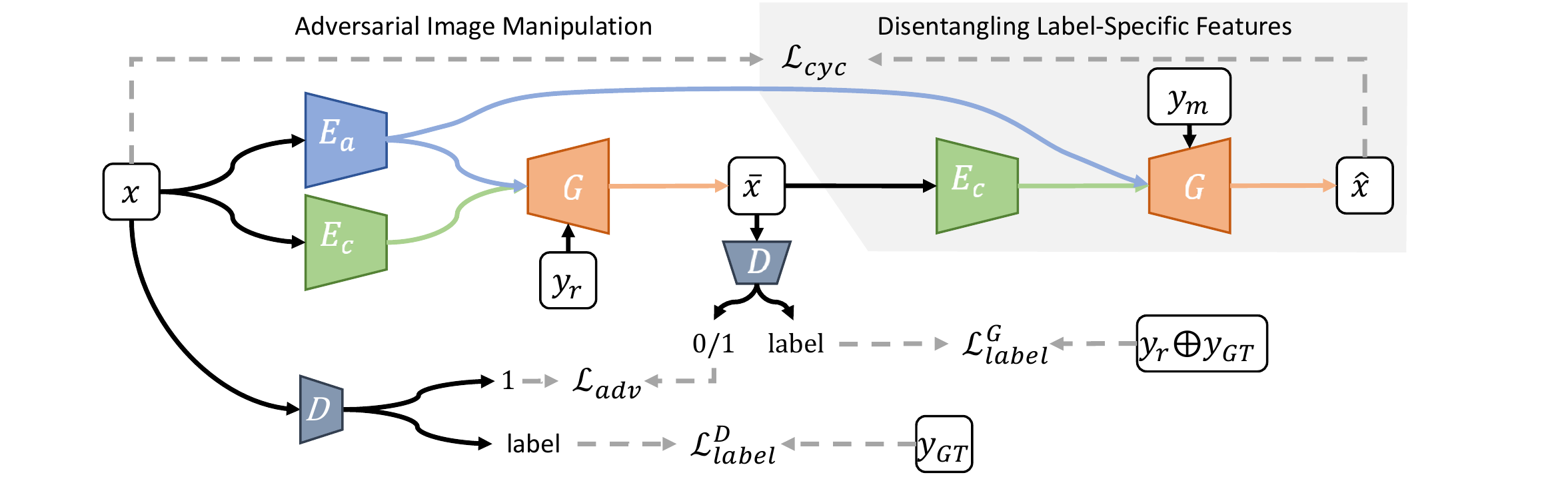}
    \end{center}
    \caption{\textbf{The schematic overview of DLGAN.} The trapezoidal blocks with the same color indicate shared parameters. The area with a grey background is for label-specific feature disentanglement, while the other area is for image manipulation. 
    }
    \label{fig:schematic}
\end{figure*}

To manipulate the image conditioned on the input labels, 
as the left of Fig.~\ref{fig:schematic} shows, 
the input image $x$ is first fed into the content encoder $E_{c}$ and the attribute encoder $E_{a}$. 
Then, the generator $G$ first receives the representations of the input image from the encoders, 
and then learns to use a random label $y_{r}$ to manipulate the input image by guiding the synthesized image $\bar{x}=G(E_c(x),E_a(x),y_r)$ to fool the discriminator $D$.
Concretely, 
the discriminator learns to classify the real image $x$ as real sample and classify the synthesized image $\bar{x}$ as fake sample. 
The generator competes with the discriminator by learning to fool the discriminator to classify the synthesized images as real samples. 

%
To stabilize the adversarial training process and generate higher-quality images, we use the Wasserstein GAN~\cite{arjovsky2017wasserstein,Gulrajani2017improved} adversarial loss with a gradient penalty as follows.




\begin{equation}
\begin{aligned}
    \mathcal{L}_{adv}= 
      & -\mathbb{E}_{x,y_r}[D_{logit}(G(E_c(x),E_a(x),y_r))]
       + \mathbb{E}_x [D_{logit}(x)] \\ 
       & - \lambda_{gp} \mathbb{E}_{\widetilde{x}}[(||\bigtriangledown_{\widetilde{x}}D_{logit}(\widetilde{x})||_{2}-1)^2] \\
\end{aligned}
\end{equation}
where $D_{logit}$ denotes the real/fake output of discriminator, $\lambda_{gp}$ is the weight of the gradient penalty loss and $\widetilde{x}$ is sampled uniformly along a straight line between a pair of a real and a synthesized image~\cite{arjovsky2017wasserstein}. 

In addition, to ensure the synthesized images match with the input label, the discriminator learns to classify the real image $x$ into the ground truth label $y_{GT}$.
Meanwhile, the generator learns to synthesize image $\bar{x}$ to match with the desired label $y_r \oplus y_{GT}$. In other words, if $y_{r}$ has missing features, the synthesized image $\bar{x}$ should maintain the corresponding original features of the input image $x$. To achieve that, two label losses are applied to the discriminator and generator, respectively:

\begin{equation}
    \mathcal{L}_{label}^{D}=\mathbb{E}_{x}[\mathcal{CE}(D_{label}(x),y_{GT})]
\end{equation}
\begin{equation}
\begin{aligned}
    \mathcal{L}_{label}^{G}=
    &\mathbb{E}_{x,y_r}[\mathcal{CE}(D_{label}(\bar{x}),y_r\oplus y_{GT})],\\
\end{aligned}
\end{equation}
where $\mathcal{CE}$ stands for cross-entropy and $D_{label}$ denotes the output label of discriminator. 

\subsection{Disentangling Label-Specific Features}
\label{ch:disentanglement}

We can now manipulate the input images using the random label $y_r$, but how to disentangle and map the label-specific features into a prior distribution?
To achieve that, as the schematic indicated by the grey background in Fig.~\ref{fig:schematic} shows, we reconstruct the input image $x$ by $\hat{x}=G(E_c(\bar{x}), E_a(x), y_m)$. The cycle-consistent loss $\mathcal{L}_{cyc}$ between the input image $x$ and reconstructed image $\hat{x}$ guarantees that the output of the attribute encoder $E_{a}$ contains the label-specific features.
The reason is that by using a random label $y_r$ to manipulate the input image $x$, the synthesized/manipulated image $\bar{x}$ does not always contain all label-specific features of the original input image $x$. At the same time, the matching label $y_m$ also does not always contain all label-specific information of the original input image. In other words, both the $E_c(\bar{x})$ and $y_m$ cannot provide all label-specific features of the input image for the $\hat{x}$. Therefore, to reconstruct the input image under different cases of $y_r$ and $y_m$, all label-specific features of the input image $x$ must be passed through the attribute encoder $E_a$. 
The cycle-consistent loss is as follows.
\begin{equation}
\begin{aligned}
    \mathcal{L}_{cyc} = 
    &\mathbb{E}_x [||\hat{x}-x||_1],\\
\end{aligned}
\end{equation}

The label representation method described in Section~\ref{ch:label} plays an important role in our method.  Specifically, the missing-feature property of $y_m$ and $y_r$ is necessary to achieve the disentanglement of label-specific representations. If $y_m$ and $y_r$ have no missing features, no label-specific feature needs to be passed through $E_a$. Then all features of the image can be passed through $E_c$,  indicating a failure in disentanglement.
In addition, the missing-feature property enables users to manipulate a specific feature of the image without being required to provide the information of other features.

\begin{figure*}[t]
    \begin{center}
        \includegraphics[scale=0.55, trim={0cm, 8cm, 0cm, 8cm}, clip]{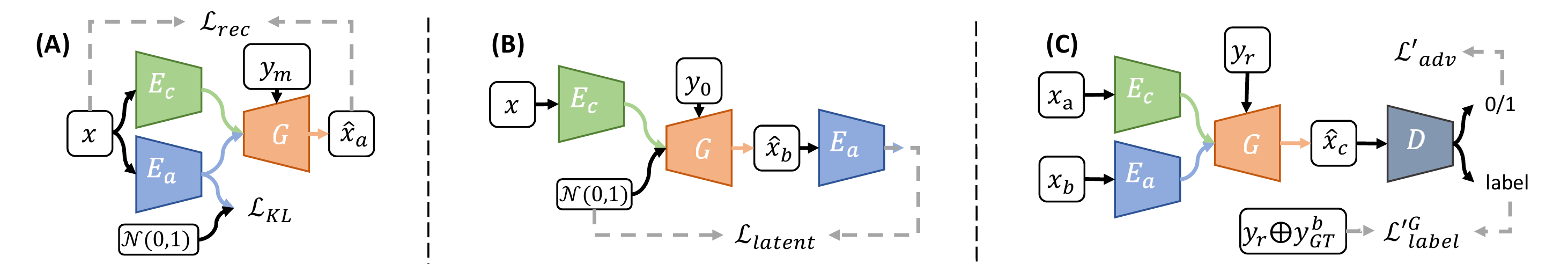}
    \end{center}
    \caption{\textbf{Other regularization losses.} 
    The identity reconstruction loss $\mathcal{L}_{rec}$ facilitates the training and encourages the generator to maintain the background of the input image better; the KL loss $\mathcal{L}_{KL}$ aims to align the attribute representation with a prior normal distribution; the latent regression loss $\mathcal{L}_{latent}$ enforces the reconstruction on the latent attribute vector; and the additional adversarial loss $\mathcal{L'}_{adv}$ and label loss $\mathcal{L'}_{label}^G$ promote further feature disentanglement. The trapezoidal blocks with the same color indicate shared parameters with Fig.~\ref{fig:schematic}.}
    \label{fig:other_reg}
\end{figure*}

Apart from the feature disentanglement, the cycle-consistent loss can ``force'' the manipulated image $\bar{x}$ to maintain information (\emph{e.g.,} background) from the input image $x$; otherwise, the reconstruction process cannot reconstruct the input images well~\cite{zhu2017cyclegan}.
%
In addition, as Part (A) of Fig.~\ref{fig:other_reg} shows, we can further apply an identity reconstruction loss $\mathcal{L}_{rec}$ to facilitate the training. Since the matching label $y_m$ can be missing in some or all label-specific features, to reconstruct $\hat{x}_a=G(E_c(x),E_a(x),y_m)$, all image features, including label-specific and content features, must pass through the $E_a$ and $E_c$. The identity reconstruction loss $\mathcal{L}_{rec}$ is as follows. 
\begin{equation}
    \mathcal{L}_{rec}=\mathbb{E}_x [||\hat{x}_a-x||_1]
\end{equation}


To support smooth manipulation, \emph{e.g.,} c hanging the hair color from one to another smoothly rather than by an immediate change, and to perform stochastic sampling, we need to map the attribute features into a prior normal distribution.
To achieve that, the output of $E_a$ is set to be a 1D vector,  
and we apply a KL-divergence loss between the output of $E_a$ and the normal distribution $\mathcal{N}(0,1)$, as shown in Part ($A$) of Fig.~\ref{fig:other_reg}. 
The KL-divergence loss $\mathcal{L}_{KL}$ is as follows.
\begin{equation}
    \mathcal{L}_{KL}=\mathbb{E}_x [\mathcal{D}_{KL}(E_a(x)||\mathcal{N}(0,1))]
\end{equation}
where $\mathcal{D}_{KL}(p||q)=-\int p(z)log\frac{p(z)}{q(z)}dz$ denotes KL-divergence between two distributions. 






\subsection{Other Regularization}
\label{ch:other_reg}

\begin{figure}[t!]
    \begin{center}
        \includegraphics[scale=0.6, trim={8cm, 6cm, 1cm, 6cm}, clip]{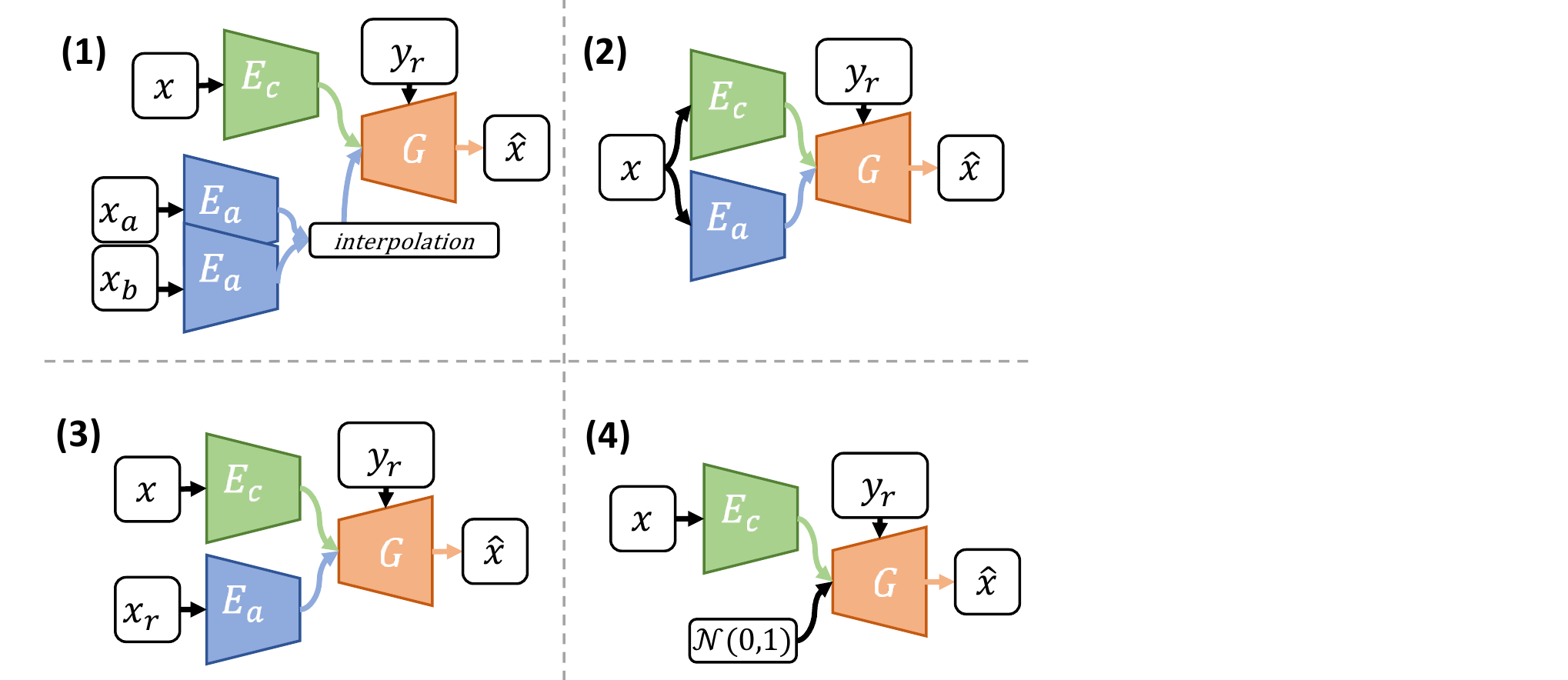}
    \end{center}
    \caption{\textbf{Inferences for different applications.} (1) Manipulate the attributes of $x$ by interpolating the attribute features of two reference images $x_a$ and $x_b$ while additional label $y_r$ can be used as a constraint. (2) Manipulate the image $x$'s attributes using label $y_r$ only. (3) Manipulate the image $x$'s attributes using both label $y_r$ and reference image $x_r$, where 
    the $y_r$ has higher priority than $x_r$. (4) Manipulate the image $x$'s attributes randomly while additional label $y_r$ can be used as a constraint.}
    \label{fig:app_illustration}
\end{figure}

\subsubsection{Latent Regression Loss}
To encourage invertible mapping between the image space
and the latent space, we can further apply an additional latent regression loss $\mathcal{L}_{latent}$~\cite{zhu2017toward}.
As shown in Part ($B$) of Fig.~\ref{fig:other_reg},
we can sample a latent vector $z$ from the prior normal distribution as the attribute representation and attempt to reconstruct $z$ with $\hat{z}=E_a(\hat{x}_b)$, where $\hat{x}_b=G(E_c(x),z,y_0)$. The latent regression loss $\mathcal{L}_{latent}$ can be formulated as follows.
\begin{equation}
    \mathcal{L}_{latent}=\mathbb{E}_{x,z}[||\hat{z}-z||_1]
\end{equation}

\subsubsection{Additional Disentanglement Loss}
In addition, Part ($C$) of Fig.~\ref{fig:other_reg} can help to further disentangle the label-specific attribute features by guiding the generator to generate an image with content features and attribute features from two different images.
The label loss for $G$ in Part ($C$) is as follows.
\begin{equation}
\begin{aligned}
    \mathcal{L'}^{G}_{label}
    = \mathbb{E}_{x_a,x_b,y_r}[\mathcal{CE}(D_{label}(\hat{x}_c),y_r\oplus y_{GT}^b)]
\end{aligned}
\end{equation}
where $\hat{x}_c=G(E_c(x_a),E_a(x_b),y_r)$, and $y_r\oplus y_{GT}^b$ denotes a random label where the missing features of $y_r$ are filled with the ground truth of the reference image $x_b$. $\mathcal{L'}_{label}^G$ can guide the generator to synthesize the images that match the attributes of the reference images and labels.
The adversarial loss in Part ($C$) is as follows.
\begin{equation}
\begin{aligned}
    \mathcal{L'}_{adv}= 
      -\mathbb{E}_{x_a,x_b,y_r} [D_{logit}(G(E_c(x_a),E_a(x_b),y_r))] 
\end{aligned}
\end{equation}
which is applied to the generator and reversely to the discriminator to encourage the generator to synthesize highly realistic images.

\begin{figure*}[t]
    \begin{center}
        \includegraphics[scale=0.45, trim={1.17cm, 4.3cm, 0.85cm, 7.5cm}, clip]{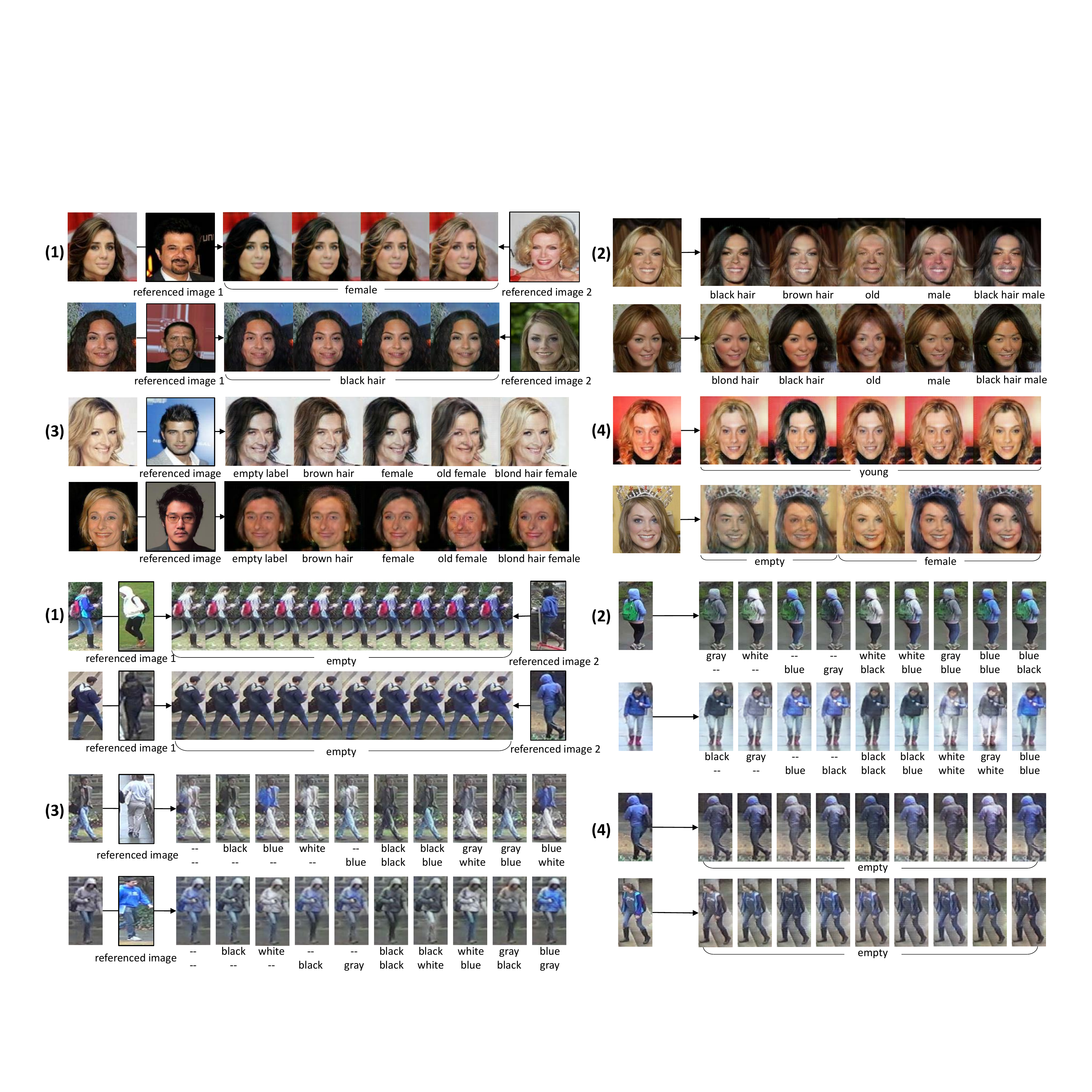}
    \end{center}
    \caption{\textbf{Example results of DLGAN on CelebA and DukeMTMC.} 
    For both datasets, we conduct experiments on all the four applications.
    For DukeMTMC, labels are shown in two rows. The upper row represents the color of upper body clothing and the lower row represents the color of lower body clothing. ``- -'' means empty label.
    More results can be seen in supplementary.}
    \label{fig:result_dlgan} 
\end{figure*}

\subsubsection{Total Loss}

To conclude, the overall loss function for generator, content encoder, and attribute encoder is as follows.
\begin{equation}
\begin{aligned}
\mathcal{L}_{Total}^{G,E_a,E_c}=&\lambda^G_{label}(\mathcal{L}^{G}_{label}+\mathcal{L'}^{G}_{label})\\
&+(\mathcal{L}_{adv}+\mathcal{L'}_{adv}) +\lambda_{cyc}\mathcal{L}_{cyc}\\
&+\lambda_{rec}\mathcal{L}_{rec}
+\lambda_{latent}\mathcal{L}_{latent} + \lambda_{KL}\mathcal{L}_{KL}
\end{aligned}
\end{equation}

The overall loss function for the discriminator is as follows.
\begin{equation}
\begin{aligned}
\mathcal{L}^{D}_{Total}=-(\mathcal{L}_{adv}+\mathcal{L'}_{adv})+\lambda^D_{label}\mathcal{L}^D_{label}
\end{aligned}
\end{equation}
where the $\lambda^G_{label}$, $\lambda_{cyc}$, $\lambda_{rec}$, $\lambda_{latent}$, $\lambda_{KL}$, and $\lambda^D_{label}$ are all scale values that control the weights of different losses.

\section{Experiment}

\subsection{Datasets and Implementation Details}

We conduct two experiments on two multi-label datasets, CelebA~\cite{liu2015faceattributes} 
and DukeMTMC~\cite{Ristani2016performance}. 
For CelebA, 
we use the labels of hair colors (black, blond, and brown), gender (male and female) and age (old and young). We use 100,000 images for training and 15,309 images for testing. 
For DukeMTMC, we choose to investigate into the colors of upper-body and lower-body clothing. We discard colors with samples fewer than 1,000, such as purple and green for upper-body and red and brown for lower-body. We use 11,643 images for training and 1,681 images for testing.

To represent the content information (\textit{e.g.,} facial appearance and background) of the input image, the output of the content encoder is a 3D tensor which can contain more spatial information than 1D vector.  
Specifically, the content encoder comprises of two convolutional layers for downsampling and three residual blocks~\cite{he2016deep}. For CelebA with the image size of $128\times 128$, the content encoder encodes images into tensors with a shape of $32\times 32\times 256$. For DukeMTMC with the image size of $64\times 128$, we use the same architecture, and the shape of content tensor becomes $16\times 32\times 256$.
The attribute encoder 
comprises of three residual blocks to encode images into vectors with a dimension of $16$.
The generator comprises of six residual blocks followed by two deconvolutional layers for upsampling.
We use the discriminator architecture of StarGAN to classify whether images are real or fake as well as classify them into their corresponding labels.
We use batch normalization for the content encoder, attribute encoder and generator but layer normalization for the discriminator.

For training,
we adopt the verified settings of $\lambda_{cyc}=10$, $\lambda_{KL}=0.01$, $\lambda^D_{label}=1$, $\lambda^G_{label}=1$ and $\lambda_{gp}=10$ according to the previous image manipulation studies~\cite{Lee2019DRIT, Choi2018starGAN}.
Besides, after trying a wide range of values for the rest of lambdas varying from 0.01 to 50, 
we empirically find that the fittest setting is $\lambda_{rec}=1$ and $\lambda_{latent}=5$. The same setting is used for both datasets.
We use Adam optimizer~\cite{KingmaAdam2014} with a batch size of 32 and an initial learning rate of 0.0001. The model is trained for 20 epochs (on CelebA) and 200 epochs (on DukeMTMC), and the learning rate is linearly decayed to zero after half of the total epoch. 
The code is publicly available at \url{http://github.com/TBA} and the training on a V100 GPU took 5 days for CelebA and 3 days for DukeMTMC. All input images are picked from
the test set for both qualitative and quantitative comparisons.

\subsection{Qualitative Evaluation}

\begin{figure*}[t]
    \begin{center}
        \includegraphics[scale=0.65, trim={0.5cm, 2cm, 1cm, 5cm}, clip]{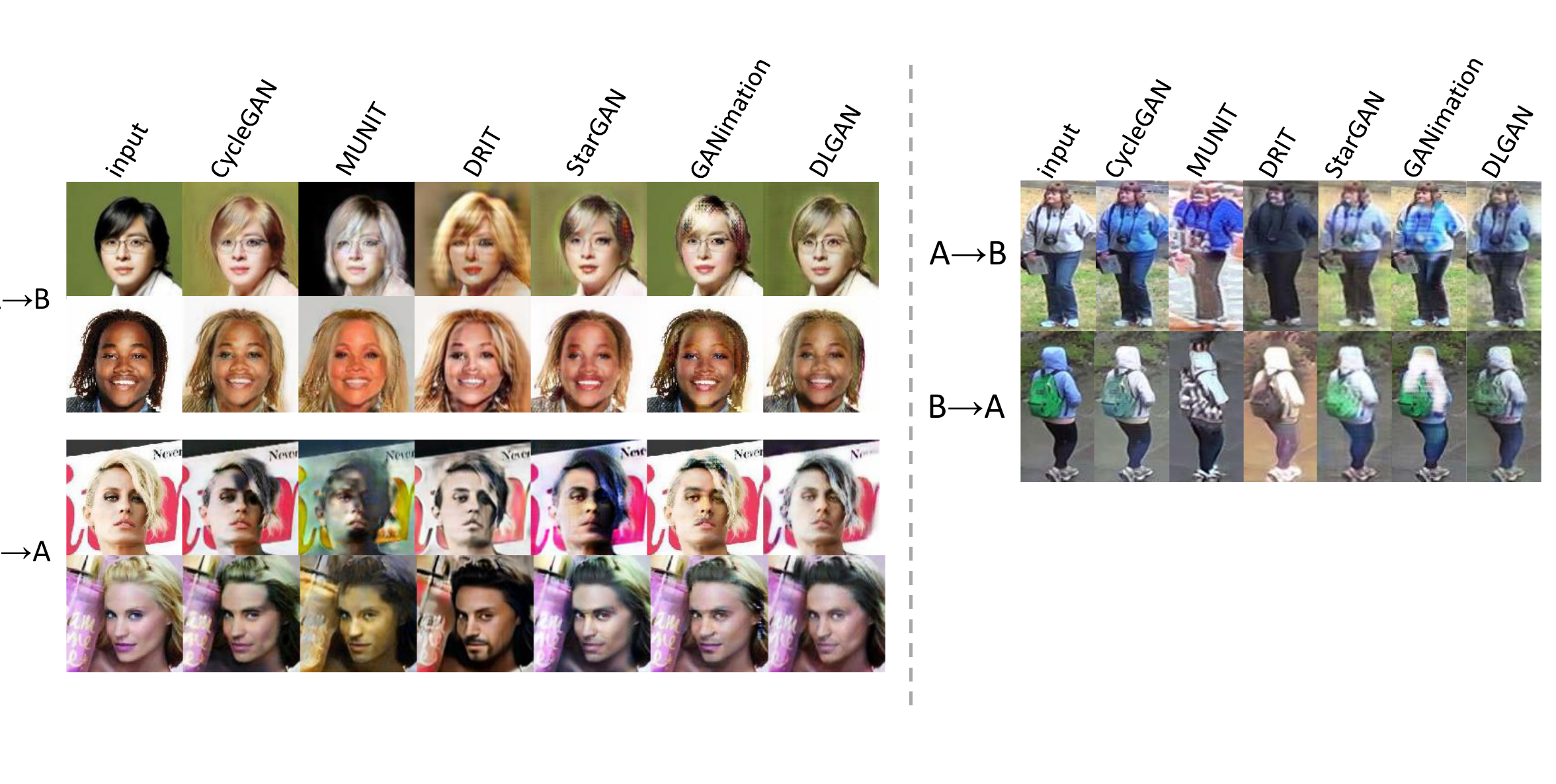}
    \end{center}
    \caption{\textbf{Comparison against baseline methods: Cross-Domain Translation.} 
    (Left) Results on CelebA, where domain A represents black hair male and domain B represents blond hair female. (Right) Results on DukeMTMC, where domain A represents blue upper body clothing, black lower body clothing and domain B represents white upper body clothing, blue lower body clothing.
    }
    \label{fig:domain_transfer}
\end{figure*}

\begin{figure*}[htb]
    \begin{center}
        \includegraphics[scale=0.7, trim={0.5cm, 2.5cm, 0cm, 2cm}, clip]{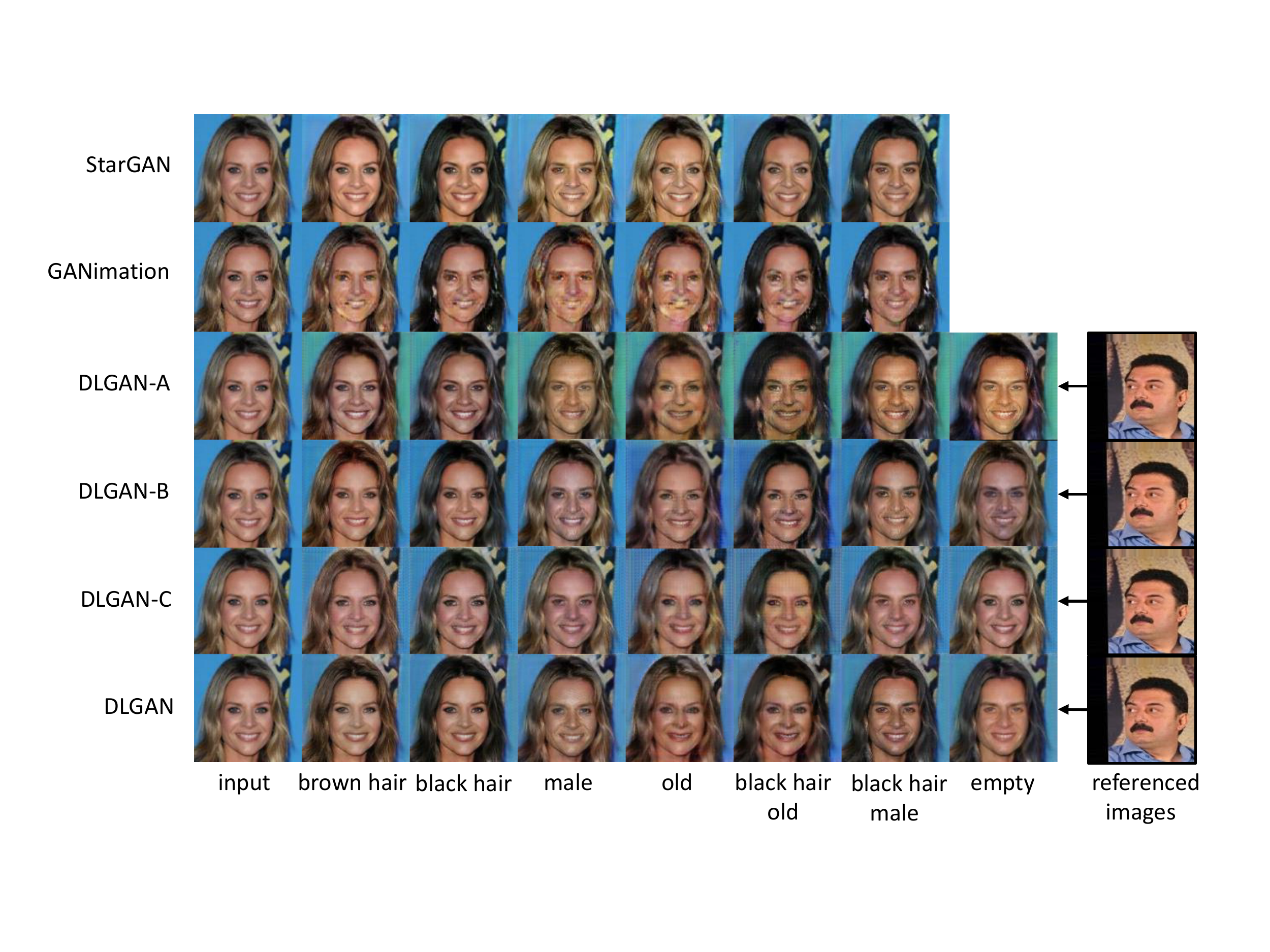}
    \end{center}
    \caption{\textbf{Qualitative comparison with baseline methods and ablation study: Multi-Domain Translation Within a Single Model.} 
    The second to the seventh columns show the translation to different domains conditioned on labels. 
    The last column shows the translation to the \emph{black hair, male} domain conditioned on reference images. 
    StarGAN and GANimation are infeasible for this task so we only show results of different settings of DLGAN. 
    }
    \label{fig:multi_domain}
\end{figure*}

\subsubsection{Four applications of DLGAN}
Fig.~\ref{fig:app_illustration} shows the schematics of different applications of DLGAN. The results on the testing set for both datasets are in Fig.~\ref{fig:result_dlgan}.
To demonstrate that DLGAN can walk on the manifold of the attribute feature space, \emph{Application $1$} uses the interpolation between the attribute features of two reference images from different domains as the new attribute representation for manipulation.
Meanwhile, a non-empty label can be used as a constraint, \emph{e.g,} as the top-left of Fig.~\ref{fig:result_dlgan} shows, the interpolated images are controlled by the given label of female.

\emph{Application $2$} provides a classical way to manipulate images conditioned on the given label only.
%
In \emph{Application $3$}, a synthesized image $\hat{x}$ is composed of the content of the input image $x$ and the attributes of the reference image $x_r$ and label $y_r$, where the label has a higher priority than the reference image.
\emph{Application $2$} can achieve Cross-Domain Translation task (\emph{i.e.,} transfer images from one domain to the other) and both \emph{Application $2$} and \emph{Application $3$} can achieve Multi-Domain Translation task (\emph{i.e.,} use a single model to translate images among multiple domains).
%
For \emph{Application $4$}, 
random noises $\mathcal{N}(0,1)$ generate a series of images $\hat{x}$ with the content identical to the input image $x$, meanwhile, an additional label $y_r$ can be used as a constraint. This application achieves multi-modal image-to-image translation.

\emph{Applications $1$} and $4$ demonstrate that DLGAN successfully maps the discrete attributes into a continuous prior distribution.
In addition, \emph{Applications $2$} and $3$ can generate images with certain attributes, which will be used for comparison against baseline methods
both qualitatively and quantitatively.

\subsubsection{Cross-Domain Translation}
 Cross-Domain Translation can be done by \emph{Application $2$} of DLGAN.
We demonstrate the effectiveness of DLGAN in Cross-Domain Translation task by comparing it with the following baseline works: 
a) CycleGAN;
b) MUNIT; 
c) DRIT;
d) StarGAN;
e) GANimation.
For fair comparison, all baseline models are trained for the same time on the same training dataset as DLGAN, and the result images are randomly picked. 
As Fig.~\ref{fig:domain_transfer} shows, DLGAN
performs competitively or better than other baselines in both visual realism and correctness of the modified attributes between two domains, as supported by the FID scores in Tables~\ref{tab:fid_celeba_domaintransfer},~\ref{tab:fid_duke_domaintransfer}, and KID scores in Tables~\ref{tab:kid_celeba_domaintransfer},~\ref{tab:kid_duke_domaintransfer}. 

\subsubsection{Multi-Domain Translation Within a Single Model}

We demonstrate the effectiveness of DLGAN in Multi-Domain Translation task by comparing it with the following baselines:
1) StarGAN;
2) GANimation;
3) without Part ($A$) of Fig.~\ref{fig:other_reg}, 
namely DLGAN-A;
4) without Part ($B$) of Fig.~\ref{fig:other_reg}, 
namely DLGAN-B;
and 
5) without Part ($C$) of Fig.~\ref{fig:other_reg}, 
namely DLGAN-C.
Since CycleGAN, MUNIT and DRIT are infeasible for multi-domain translation within a single model, we don't put their results here.
Fig.~\ref{fig:multi_domain} shows comparison among different methods; supported by the FID scores in Table~\ref{tab:fid_multi_labels}, KID scores in Table~\ref{tab:kid_multi_labels} and human evaluation in Table~\ref{table:human_eval}, DLGAN outperforms GANimation and StarGAN. 
Besides, DLGAN is able to support more applications and more challenging tasks as mentioned in Table~\ref{table:novelty}.

In addition,
DLGAN-A has poorer performance than DLGAN in terms of maintaining the background of the input images.
In some cases, the synthesized images of DLGAN-B and DLGAN-C cannot contain the attributes from the reference images $x_r$ in \emph{Application $3$}, \emph{e.g.,} as shown in the last column of DLGAN-B and DLGAN-C part of Fig.~\ref{fig:multi_domain} where the black hair attribute cannot be passed to the synthesized images when input label is missing in hair color. The results indicate that DLGAN-B and DLGAN-C have a poor performance in feature disentanglement. The ablation study helps us to analyze the effectiveness of different modules of our method. We can see that the identity reconstruction in Part ($A$) plays an important role in maintaining the background of the input images while the latent regression loss in Part ($B$) and the additional disentanglement loss in Part ($C$) are crucial to the disentanglement of label-specific features, without which the attribute features of reference images are not possible to be passed to the synthesized images.

\subsection{Quantitative Evaluation}
\label{ch:quantitative}

\subsubsection{Fr\'echet Inception Distance}

(FID) is a standard metric to evaluate the distance between two distributions, has been widely used to 
evaluate the quality of synthesized images~\cite{xiao2018,brock2019two}.
For Cross-Domain Translation, different methods manipulate all testing set images in domain A and domain
B. For Multi-Domain Translation, different methods manipulate all testing set images conditioned on the
same random discrete labels for the translation that only conditioned on discrete labels (i.e. Application 2 of
DLGAN). For the translation that conditioned on both reference images and discrete labels (i.e. Application
3 of DLGAN), different methods manipulate all testing set images conditioned on the same random discrete
labels and reference images.

Tables~\ref{tab:fid_celeba_domaintransfer},~\ref{tab:fid_multi_labels},~\ref{tab:fid_duke_domaintransfer} and~\ref{tab:fid_multi_images} list the FID scores between the distribution of real images and generated images by
different methods in both Cross-Domain Translation and Multi-Domain Translation tasks.
DLGAN outperforms other state-of-the-art methods on both datasets.
Based on these scores, we can also see that DLGAN performs better than other DLGAN settings except for DLGAN-C on DukeMTMC dataset, which shows a competitive performance with DLGAN.

\subsubsection{Kernel Inception Distance} 
(KID) measures the dissimilarity between two distributions using samples drawn independently from each distribution. It has also been widely used to evaluate the quality of the synthesized images~\cite{Ori2019breaking,Kim2020U-GAT-IT}.
For fair comparison, the same principles of comparison are adopted as mentioned in the FID experiments.
Tables~\ref{tab:kid_celeba_domaintransfer},~\ref{tab:kid_multi_labels},~\ref{tab:kid_duke_domaintransfer}, and~\ref{tab:kid_multi_images} list the KID scores 
under different methods in both Cross-Domain Translation and Multi-Domain Translation tasks.
The results are similar to those of FID, which illustrate the effectiveness of DLGAN against all baseline methods as well as ablation settings.




\begin{table*}[t!]
\begin{floatrow}
\capbtabbox{
\begin{tabular}{|c|c|c|}
\hline
\multirow{2}{*}{\textbf{\begin{tabular}[c]{@{}c@{}}Appli-\\ cation2\end{tabular}}} & \multicolumn{2}{c|}{\textbf{CelebA}} \\ \cline{2-3} 
                                                                                   & A2B               & B2A              \\ \hline
DLGAN                                                                              & \textbf{15.46}    & \textbf{30.80}   \\ \hline
DRIT                                                                               & 34.77             & 56.09            \\ \hline
MUNIT                                                                              & 23.56             & 40.06            \\ \hline
CycleGAN                                                                           & 37.84             & 42.05            \\ \hline
StarGAN                                                                            & 36.65             & 45.65            \\ \hline
GANimation                                                                         & 71.93             & 57.24            \\ \hline
\end{tabular}
}{
 \caption{FID scores for Cross-Domain Translation on CelebA.}
 \label{tab:fid_celeba_domaintransfer}
}
\capbtabbox{
\begin{tabular}{|c|c|c|c|c|c|}
\hline
\multirow{2}{*}{\textbf{\begin{tabular}[c]{@{}c@{}}Appli-\\ cation2\end{tabular}}} & \multicolumn{3}{c|}{\textbf{CelebA}}             & \multicolumn{2}{c|}{\textbf{DukeMTMC}} \\ \cline{2-6} 
                                                                                   & hair           & gender         & age            & down               & up                \\ \hline
DLGAN                                                                              & \textbf{10.60} & \textbf{14.28} & \textbf{13.73} & \textbf{84.94}     & \textbf{91.91}    \\ \hline
DLGAN-A                                                                            & 32.36          & 38.96          & 34.71          & 101.98             & 106.51            \\ \hline
DLGAN-B                                                                            & 38.01          & 43.82          & 43.15          & 115.35             & 119.05            \\ \hline
DLGAN-C                                                                            & 16.50          & 26.26          & 19.88          & 97.74              & 105.77            \\ \hline
StarGAN                                                                            & 12.69          & 18.12          & 14.02          & 101.83             & 105.95            \\ \hline
GANimation                                                                         & 12.28          & 16.76          & 13.87          & 92.62              & 93.79             \\ \hline
\end{tabular}
}{
 \caption{FID scores for Multi-Domain Translation conditioned on discrete labels.}
 \label{tab:fid_multi_labels}
}
\end{floatrow}
\begin{floatrow}
\capbtabbox{
\begin{tabular}{|c|c|c|}
\hline
\multirow{2}{*}{\textbf{\begin{tabular}[c]{@{}c@{}}Appli-\\ cation2\end{tabular}}} & \multicolumn{2}{c|}{\textbf{DukeMTMC}} \\ \cline{2-3} 
                                                                                   & A2B                & B2A               \\ \hline
DLGAN                                                                              & \textbf{194.32}    & \textbf{178.29}   \\ \hline
DRIT                                                                               & 238.25             & 270.36            \\ \hline
MUNIT                                                                              & 224.68             & 214.46            \\ \hline
CycleGAN                                                                           & 197.98             & 180.84            \\ \hline
StarGAN                                                                            & 208.83             & 200.34            \\ \hline
GANimation                                                                         & 207.93             & 211.00            \\ \hline
\end{tabular}
}{
 \caption{FID scores for Cross-Domain Translation on DukeMTMC.}
 \label{tab:fid_duke_domaintransfer}
}
\capbtabbox{
\begin{tabular}{|c|c|c|c|c|c|}
\hline
\multirow{2}{*}{\textbf{\begin{tabular}[c]{@{}c@{}}Appli-\\ cation3\end{tabular}}} & \multicolumn{3}{c|}{\textbf{CelebA}}             & \multicolumn{2}{c|}{\textbf{DukeMTMC}} \\ \cline{2-6} 
                                                                                   & hair           & gender         & age            & down              & up                 \\ \hline
DLGAN                                                                              & \textbf{21.56} & \textbf{30.67} & \textbf{26.31} & \textbf{93.08}    & 116.39             \\ \hline
DLGAN-A                                                                            & 34.63          & 40.29          & 38.91          & 131.22            & 149.28             \\ \hline
DLGAN-B                                                                            & 50.81          & 53.89          & 54.57          & 104.53            & 121.77             \\ \hline
DLGAN-C                                                                            & 36.06          & 42.85          & 38.46          & 95.50             & \textbf{115.76}    \\ \hline
\end{tabular}
}{
 \caption{FID scores for Multi-Domain Translation conditioned on reference images and discrete labels.}
 \label{tab:fid_multi_images}
}
\end{floatrow}
\end{table*}

\begin{table*}[t!]
\begin{floatrow}
\capbtabbox{
\begin{tabular}{|c|c|c|}
\hline
\multirow{2}{*}{\textbf{\begin{tabular}[c]{@{}c@{}}Appli-\\ cation2\end{tabular}}} & \multicolumn{2}{c|}{\textbf{CelebA}}                                                                                              \\ \cline{2-3} 
                                                                                   & A2B                                                             & B2A                                                             \\ \hline
DLGAN                                                                              & \textbf{\begin{tabular}[c]{@{}c@{}}0.008\\ \tiny{$\pm 0.0003$}\end{tabular}} & \textbf{\begin{tabular}[c]{@{}c@{}}0.022\\ \tiny{$\pm 0.0009$}\end{tabular}} \\ \hline
DRIT                                                                               & \begin{tabular}[c]{@{}c@{}}0.034\\ \tiny{$\pm 0.0004$}\end{tabular}          & \begin{tabular}[c]{@{}c@{}}0.055\\ \tiny{$\pm 0.0005$}\end{tabular}          \\ \hline
MUNIT                                                                              & \begin{tabular}[c]{@{}c@{}}0.023\\ \tiny{$\pm 0.0004$}\end{tabular}          & \begin{tabular}[c]{@{}c@{}}0.035\\ \tiny{$\pm 0.0005$}\end{tabular}          \\ \hline
CycleGAN                                                                           & \begin{tabular}[c]{@{}c@{}}0.042\\ \tiny{$\pm 0.0006$}\end{tabular}          & \begin{tabular}[c]{@{}c@{}}0.040\\ \tiny{$\pm 0.0006$}\end{tabular}          \\ \hline
StarGAN                                                                            & \begin{tabular}[c]{@{}c@{}}0.037\\ \tiny{$\pm 0.0007$}\end{tabular}          & \begin{tabular}[c]{@{}c@{}}0.040\\ \tiny{$\pm 0.0008$}\end{tabular}          \\ \hline
GANimation                                                                         & \begin{tabular}[c]{@{}c@{}}0.082\\ \tiny{$\pm 0.0010$}\end{tabular}          & \begin{tabular}[c]{@{}c@{}}0.054\\ \tiny{$\pm 0.0009$}\end{tabular}          \\ \hline
\end{tabular}
}{
 \caption{KID scores for Cross-Domain Translation on CelebA.}
 \label{tab:kid_celeba_domaintransfer}
}
\capbtabbox{
\begin{tabular}{|c|c|c|c|c|c|}
\hline
\multirow{2}{*}{\textbf{\begin{tabular}[c]{@{}c@{}}Appli-\\ cation2\end{tabular}}} & \multicolumn{3}{c|}{\textbf{CelebA}}                                                                                                                                                                & \multicolumn{2}{c|}{\textbf{DukeMTMC}}                                                                                            \\ \cline{2-6} 
                                                                                   & hair                                                            & gender                                                          & age                                                             & down                                                            & up                                                              \\ \hline
DLGAN                                                                              & \textbf{\begin{tabular}[c]{@{}c@{}}0.007\\ \tiny{$\pm 0.0004$}\end{tabular}} & \textbf{\begin{tabular}[c]{@{}c@{}}0.011\\ \tiny{$\pm 0.0004$}\end{tabular}} & \textbf{\begin{tabular}[c]{@{}c@{}}0.009\\ \tiny{$\pm 0.0003$}\end{tabular}} & \textbf{\begin{tabular}[c]{@{}c@{}}0.023\\ \tiny{$\pm 0.0021$}\end{tabular}} & \textbf{\begin{tabular}[c]{@{}c@{}}0.026\\ \tiny{$\pm 0.0021$}\end{tabular}} \\ \hline
DLGAN-A                                                                            & \begin{tabular}[c]{@{}c@{}}0.024\\ \tiny{$\pm 0.0005$}\end{tabular}          & \begin{tabular}[c]{@{}c@{}}0.027\\ \tiny{$\pm 0.0005$}\end{tabular}          & \begin{tabular}[c]{@{}c@{}}0.027\\ \tiny{$\pm 0.0005$}\end{tabular}          & \begin{tabular}[c]{@{}c@{}}0.040\\ \tiny{$\pm 0.0025$}\end{tabular}          & \begin{tabular}[c]{@{}c@{}}0.046\\ \tiny{$\pm 0.0026$}\end{tabular}          \\ \hline
DLGAN-B                                                                            & \begin{tabular}[c]{@{}c@{}}0.029\\ \tiny{$\pm 0.0006$}\end{tabular}          & \begin{tabular}[c]{@{}c@{}}0.034\\ \tiny{$\pm 0.0008$}\end{tabular}          & \begin{tabular}[c]{@{}c@{}}0.032\\ \tiny{$\pm 0.0006$}\end{tabular}          & \begin{tabular}[c]{@{}c@{}}0.050\\ \tiny{$\pm 0.0032$}\end{tabular}          & \begin{tabular}[c]{@{}c@{}}0.055\\ \tiny{$\pm 0.0031$}\end{tabular}          \\ \hline
DLGAN-C                                                                            & \begin{tabular}[c]{@{}c@{}}0.019\\ \tiny{$\pm 0.0004$}\end{tabular}          & \begin{tabular}[c]{@{}c@{}}0.022\\ \tiny{$\pm 0.0004$}\end{tabular}          & \begin{tabular}[c]{@{}c@{}}0.023\\ \tiny{$\pm 0.0004$}\end{tabular}          & \begin{tabular}[c]{@{}c@{}}0.033\\ \tiny{$\pm 0.0021$}\end{tabular}          & \begin{tabular}[c]{@{}c@{}}0.037\\ \tiny{$\pm 0.0022$}\end{tabular}          \\ \hline
StarGAN                                                                            & \begin{tabular}[c]{@{}c@{}}0.012\\ \tiny{$\pm 0.0004$}\end{tabular}          & \begin{tabular}[c]{@{}c@{}}0.016\\ \tiny{$\pm 0.0004$}\end{tabular}          & \begin{tabular}[c]{@{}c@{}}0.015\\ \tiny{$\pm 0.0004$}\end{tabular}          & \begin{tabular}[c]{@{}c@{}}0.039\\ \tiny{$\pm 0.0026$}\end{tabular}          & \begin{tabular}[c]{@{}c@{}}0.044\\ \tiny{$\pm 0.0026$}\end{tabular}          \\ \hline
GANimation                                                                         & \begin{tabular}[c]{@{}c@{}}0.010\\ \tiny{$\pm 0.0004$}\end{tabular}          & \begin{tabular}[c]{@{}c@{}}0.015\\ \tiny{$\pm 0.0004$}\end{tabular}          & \begin{tabular}[c]{@{}c@{}}0.012\\ \tiny{$\pm 0.0003$}\end{tabular}          & \begin{tabular}[c]{@{}c@{}}0.029\\ \tiny{$\pm 0.0021$}\end{tabular}          & \begin{tabular}[c]{@{}c@{}}0.031\\ \tiny{$\pm 0.0020$}\end{tabular}          \\ \hline
\end{tabular}
}{
 \caption{KID scores for Multi-Domain Translation conditioned on discrete labels.}
 \label{tab:kid_multi_labels}
}
\end{floatrow}
\begin{floatrow}
\capbtabbox{
\begin{tabular}{|c|c|c|}
\hline
\multirow{2}{*}{\textbf{\begin{tabular}[c]{@{}c@{}}Appli-\\ cation2\end{tabular}}} & \multicolumn{2}{c|}{\textbf{DukeMTMC}}                                                                                            \\ \cline{2-3} 
                                                                                   & A2B                                                             & B2A                                                             \\ \hline
DLGAN                                                                              & \textbf{\begin{tabular}[c]{@{}c@{}}0.050\\ \tiny{$\pm 0.0072$}\end{tabular}} & \textbf{\begin{tabular}[c]{@{}c@{}}0.038\\ \tiny{$\pm 0.0061$}\end{tabular}} \\ \hline
DRIT                                                                               & \begin{tabular}[c]{@{}c@{}}0.111\\ \tiny{$\pm 0.0064$}\end{tabular}          & \begin{tabular}[c]{@{}c@{}}0.186\\ \tiny{$\pm 0.0066$}\end{tabular}          \\ \hline
MUNIT                                                                              & \begin{tabular}[c]{@{}c@{}}0.062\\ \tiny{$\pm 0.0085$}\end{tabular}          & \begin{tabular}[c]{@{}c@{}}0.051\\ \tiny{$\pm 0.0077$}\end{tabular}          \\ \hline
CycleGAN                                                                           & \begin{tabular}[c]{@{}c@{}}0.059\\ \tiny{$\pm 0.0111$}\end{tabular}          & \begin{tabular}[c]{@{}c@{}}0.047\\ \tiny{$\pm 0.0063$}\end{tabular}          \\ \hline
StarGAN                                                                            & \begin{tabular}[c]{@{}c@{}}0.067\\ \tiny{$\pm 0.0064$}\end{tabular}          & \begin{tabular}[c]{@{}c@{}}0.053\\ \tiny{$\pm 0.0102$}\end{tabular}          \\ \hline
GANimation                                                                         & \begin{tabular}[c]{@{}c@{}}0.059\\ \tiny{$\pm 0.0098$}\end{tabular}          & \begin{tabular}[c]{@{}c@{}}0.058\\ \tiny{$\pm 0.0077$}\end{tabular}          \\ \hline
\end{tabular}
}{
 \caption{KID scores for Cross-Domain Translation on DukeMTMC.}
 \label{tab:kid_duke_domaintransfer}
}
\capbtabbox{
\begin{tabular}{|c|c|c|c|c|c|}
\hline
\multirow{2}{*}{\textbf{\begin{tabular}[c]{@{}c@{}}Appli-\\ cation3\end{tabular}}} & \multicolumn{3}{c|}{\textbf{CelebA}}                                                                                                                                     & \multicolumn{2}{c|}{\textbf{DukeMTMC}}                                                                          \\ \cline{2-6} 
                                                                                   & hair                                                   & gender                                                 & age                                                    & down                                                   & up                                                     \\ \hline
DLGAN                                                                              & \begin{tabular}[c]{@{}c@{}}\textbf{0.017}\\ \tiny{$\pm$ 0.0004}\end{tabular} & \begin{tabular}[c]{@{}c@{}}\textbf{0.027}\\ \tiny{$\pm$ 0.0006}\end{tabular} & \begin{tabular}[c]{@{}c@{}}\textbf{0.020}\\ \tiny{$\pm$ 0.0004}\end{tabular} & \begin{tabular}[c]{@{}c@{}}\textbf{0.025}\\ \tiny{$\pm$ 0.0022}\end{tabular} & \begin{tabular}[c]{@{}c@{}}\textbf{0.028}\\ \tiny{$\pm$ 0.0024}\end{tabular} \\ \hline
DLGAN-A                                                                            & \begin{tabular}[c]{@{}c@{}}0.028\\ \tiny{$\pm 0.0005$}\end{tabular} & \begin{tabular}[c]{@{}c@{}}0.033\\ \tiny{$\pm 0.0006$}\end{tabular} & \begin{tabular}[c]{@{}c@{}}0.032\\ \tiny{$\pm 0.0005$}\end{tabular} & \begin{tabular}[c]{@{}c@{}}0.056\\ \tiny{$\pm 0.0028$}\end{tabular} & \begin{tabular}[c]{@{}c@{}}0.066\\ \tiny{$\pm 0.0030$}\end{tabular} \\ \hline
DLGAN-B                                                                            & \begin{tabular}[c]{@{}c@{}}0.031\\ \tiny{$\pm 0.0004$}\end{tabular} & \begin{tabular}[c]{@{}c@{}}0.035\\ \tiny{$\pm 0.0006$}\end{tabular} & \begin{tabular}[c]{@{}c@{}}0.036\\ \tiny{$\pm 0.0004$}\end{tabular} & \begin{tabular}[c]{@{}c@{}}0.035\\ \tiny{$\pm 0.0021$}\end{tabular} & \begin{tabular}[c]{@{}c@{}}0.029\\ \tiny{$\pm 0.0028$}\end{tabular} \\ \hline
DLGAN-C                                                                            & \begin{tabular}[c]{@{}c@{}}0.027\\ \tiny{$\pm 0.0006$}\end{tabular} & \begin{tabular}[c]{@{}c@{}}0.031\\ \tiny{$\pm 0.0007$}\end{tabular} & \begin{tabular}[c]{@{}c@{}}0.033\\ \tiny{$\pm 0.0006$}\end{tabular} & \begin{tabular}[c]{@{}c@{}}0.027\\ \tiny{$\pm 0.0020$}\end{tabular} & \begin{tabular}[c]{@{}c@{}}0.030\\ \tiny{$\pm 0.0025$}\end{tabular} \\ \hline
\end{tabular}
}{
 \caption{KID scores for Multi-Domain Translation conditioned on reference images and discrete labels.}
 \label{tab:kid_multi_images}
}
\end{floatrow}
\end{table*}

\subsubsection{Human Evaluation}

To further evaluate our method, similar to previous studies~\cite{Choi2018starGAN, zhu2017cyclegan}, we use Amazon Mechanical Turk (AMT) to compare the perceptual visual fidelity (\emph{i.e.,} realism) and the correctness of the modified attribute features of our method against different baseline models. 
Specifically, we present the AMT workers with image pairs side by side on the screen. 
For each pair, one image is generated by DLGAN while the other image is generated by a randomly picked baseline model.
To make the workers not know from which model the images are generated, the image positions are random in the pair, \emph{i.e.,} the image generated by DLGAN on the left or right is of equal possibility.
%
%

We give unlimited time to the workers to make the selection. For fair comparison, we randomly generate 2,000 image pairs for CelebA and 1,000 image pairs for DukeMTMC conditioned on randomly picked labels and reference images. 
For quality control, each image pair is compared by 5 different workers, and 
we only approve workers with a life-time task approval rate greater than 98\% to participate in the evaluation. 
The workers are required to compare the image pairs based on the following criteria: a) Whether the synthesized image is realistic as well as maintaining the background and facial appearance in the original image. b) Whether the modified attributes are correct.


%
As shown in Figure~\ref{fig:app_illustration}, for \emph{Application $2$}, the criterion for correctness is that the modified image should correspond to the label $y_r\oplus y_{GT}$.
For \emph{Application $3$}, the criterion for correctness is almost the same with the only difference that the target attributes become $y_r\oplus y_{GT}^b$, where the $y_{GT}^b$ is from the reference image.

Table~\ref{table:human_eval} shows the human evaluation results. The numbers in the table indicate the ratio of workers who favor the results of DLGAN to those who favor the results of competing methods. The ratio greater than one means that DLGAN has better performance than the competing method.
Column ``\textbf{a}'' and ``\textbf{b}'' are the evaluations of realism and correctness of the modified attributes, respectively.

To summarize, the results of human evaluation conform to those of the machine evaluation explained above. As can be seen, all numbers in the table are greater than 1, indicating that DLGAN outperforms other baseline settings in both \emph{Application 2} and \emph{Application 3} for Multi-Domain Translation tasks.

\begin{table*}[t]
\centering
\begin{tabular}{|c|c|c|c|c|lccccc}
\cline{1-5}
\multirow{2}{*}{\textbf{\begin{tabular}[c]{@{}c@{}}Appli-\\ cation2\end{tabular}}} & \multicolumn{2}{r|}{\textbf{CelebA}} & \multicolumn{2}{r|}{\textbf{DukeMTMC}} &                       & \multirow{2}{*}{\textbf{}}                                                                              & \multicolumn{2}{c}{\textbf{}}                          & \multicolumn{2}{c}{\textbf{}}                         \\ \cline{2-5}
                                                                                   & a                 & b                & a                  & b                 &                       &                                                                                                         &                            &                           &                           &                           \\ \cline{1-5} \cline{7-11} 
DLGAN-A                                                                            & 1.86              & 1.33             & 1.41               & 1.89              & \multicolumn{1}{l|}{} & \multicolumn{1}{c|}{\multirow{2}{*}{\textbf{\begin{tabular}[c]{@{}c@{}}Appli-\\ cation3\end{tabular}}}} & \multicolumn{2}{c|}{\textbf{CelebA}}                   & \multicolumn{2}{c|}{\textbf{DukeMTMC}}                \\ \cline{1-5} \cline{8-11} 
DLGAN-B                                                                            & 2.19              & 2.04             & 1.83               & 2.65              & \multicolumn{1}{l|}{} & \multicolumn{1}{c|}{}                                                                                   & \multicolumn{1}{c|}{a}     & \multicolumn{1}{c|}{b}    & \multicolumn{1}{c|}{a}    & \multicolumn{1}{c|}{b}    \\ \cline{1-5} \cline{7-11} 
DLGAN-C                                                                            & 1.14              & 1.07             & 1.41               & 1.76              & \multicolumn{1}{l|}{} & \multicolumn{1}{c|}{DLGAN-A}                                                                            & \multicolumn{1}{c|}{2.59}  & \multicolumn{1}{c|}{1.63} & \multicolumn{1}{c|}{3.17} & \multicolumn{1}{c|}{2.98} \\ \cline{1-5} \cline{7-11} 
StarGAN                                                                            & 1.15              & 1.04             & 1.16               & 1.23              & \multicolumn{1}{l|}{} & \multicolumn{1}{c|}{DLGAN-B}                                                                            & \multicolumn{1}{c|}{13.10} & \multicolumn{1}{c|}{2.52} & \multicolumn{1}{c|}{2.24} & \multicolumn{1}{c|}{3.11} \\ \cline{1-5} \cline{7-11} 
GANimation                                                                         & 1.13              & 1.02             & 1.10               & 1.19              & \multicolumn{1}{l|}{} & \multicolumn{1}{c|}{DLGAN-C}                                                                            & \multicolumn{1}{c|}{2.34}  & \multicolumn{1}{c|}{2.01} & \multicolumn{1}{c|}{1.52} & \multicolumn{1}{c|}{2.61} \\ \cline{1-5} \cline{7-11} 
\end{tabular}
 \caption{\textbf{AMT perceptual evaluation of different methods in Multi-Domain Translation tasks.} 
 The numbers indicate the ratio of users who favor the results of DLGAN over users who favor a certain competing method. The ratio greater than one means that DLGAN has better performance than the competing method.
     }
 \label{table:human_eval}
\end{table*}

\section{Conclusions}
This paper proposes a framework, dubbed DLGAN, for disentangling the discrete fine-grained features of images, and enabling
smooth interpolation between different domains without continuous labels as the supervision. 
Our method also supports a hybrid manipulation using both labels and reference images as the conditions, thus a more controllable image manipulation.
%
Even though we have demonstrated the effectiveness of our method on CelebA and DukeMTMC datasets, our method is nevertheless unable to support geometric manipulation, which is useful for some image editing, \emph{e.g.,} hair style changes.
In the future work, we are interested in studying how to support image translation with large geometric changes.

\bibliographystyle{splncs04}
\bibliography{reference}

\appendices
\section{Additional Mathematical Derivation}

\begin{figure}[tb]
    \begin{center}
        \includegraphics[scale=0.4, trim={6.5cm, 5cm, 6.5cm, 4cm}, clip]{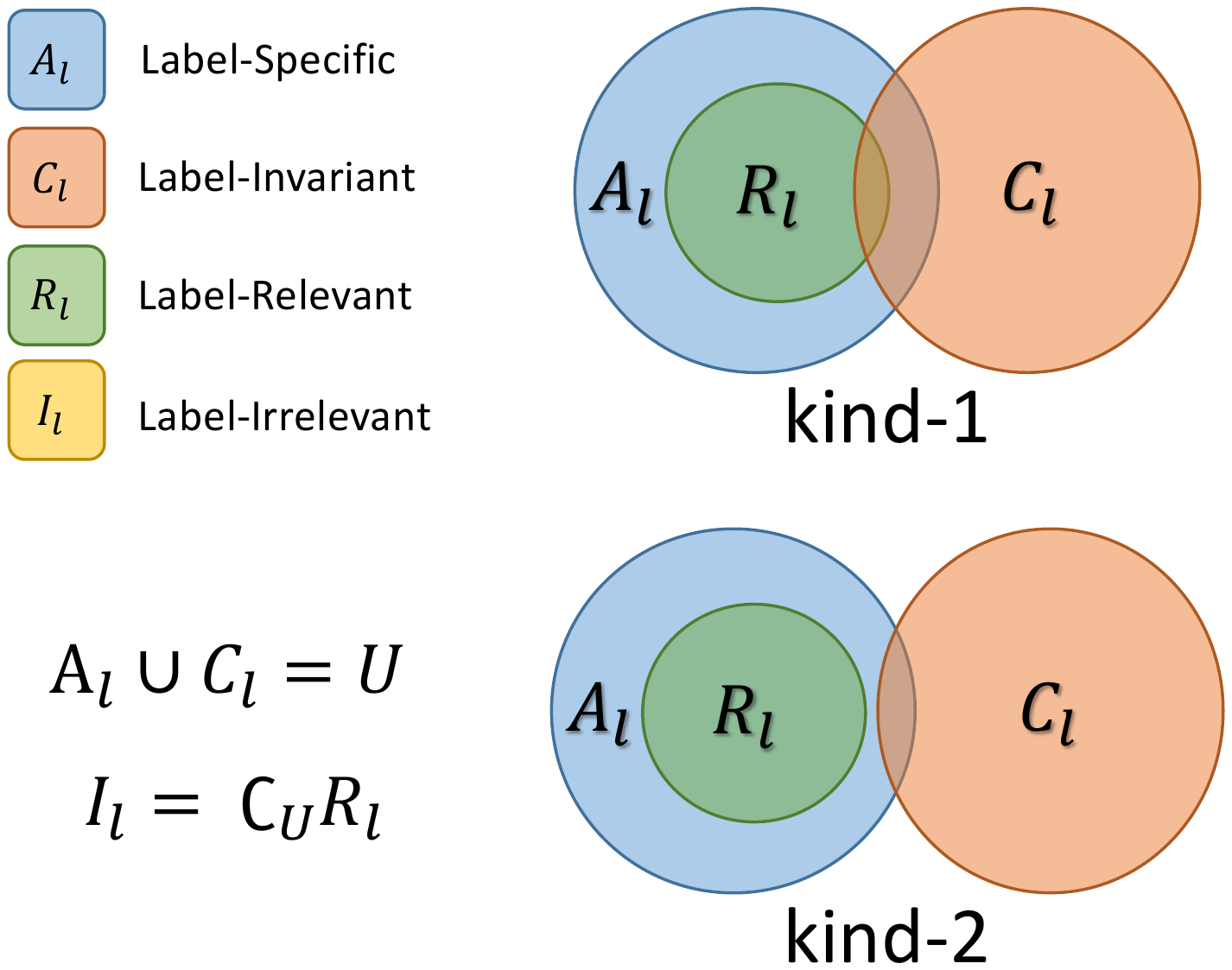}
    \end{center}
    \caption{\textbf{Venn diagram of different feature space.} $A_l$ denotes label-specific attribute space, $C_l$ denotes content space, $R_l$ denotes label-relevant space, $I_l$ denotes label-irrelevant space ($I_l= \complement_{U} R_l$ does not appear in the diagram). There are two kinds of relationship of these different feature spaces. Part ($B$) of Figure~\ref{fig:other_reg} encourages the translation from kind-1 to kind-2.}
    \label{fig:relationship}
\end{figure}

\begin{figure*}[htb]
    \begin{center}
        \includegraphics[scale=1.2, trim={3.7cm, 4.5cm, 3cm, 10.1cm}, clip]{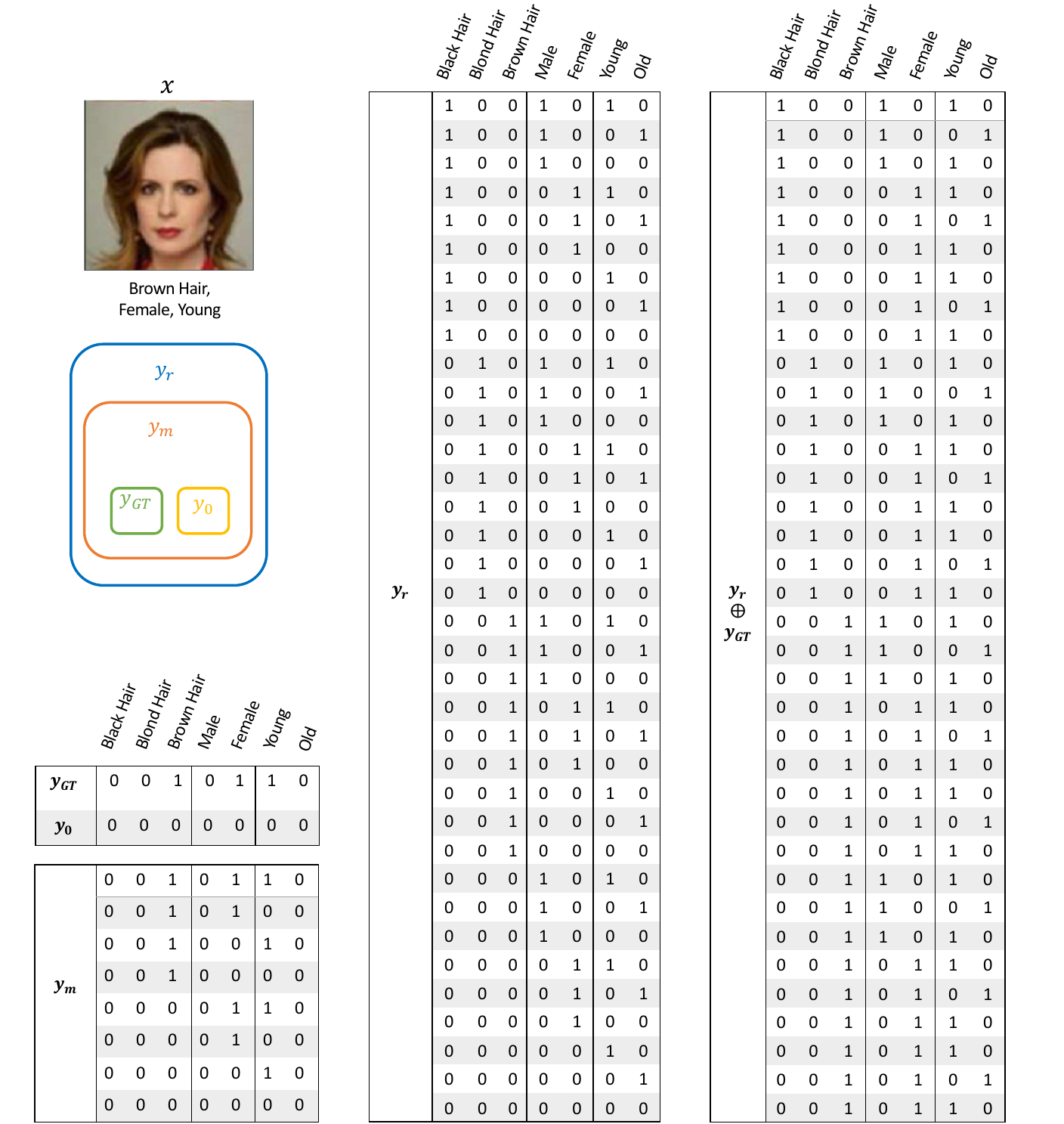}
        \caption{Different types of labels of a face image of a young female with brown hair.
        }
        \label{fig:sup_label}
    \end{center}
\end{figure*}

\begin{figure*}[htb]
    \begin{center}
        \includegraphics[scale=0.9, trim={0cm, 6cm, 0cm, 5cm}, clip]{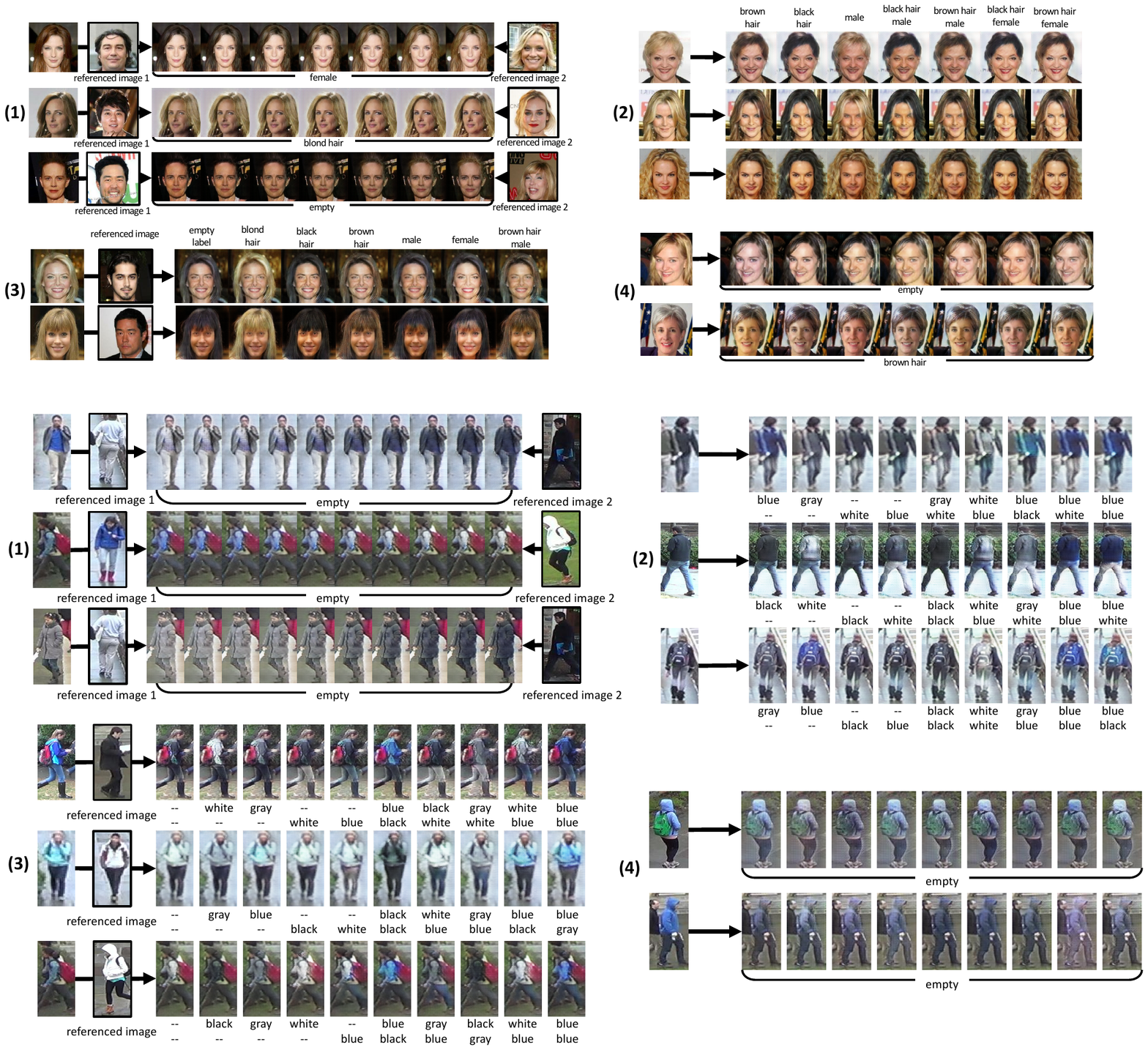}
        \caption{Additional results on CelebA dataset~\cite{liu2015faceattributes} and DukeMTMC dataset~\cite{Ristani2016performance}.}
        \label{fig:sup2}
    \end{center}
\end{figure*}

Our method can be explained using set theory.
Let $\mathcal{X} \subset \mathbb{R}^{H\times W\times 3}$ be the space of a certain domain of images, $x\in \mathcal{X}$ be a certain sample image in this domain. $L = \{l|f_i\in l \iff M_i=1, M\in \{0,1\}^n, i\in [n] \}$ denotes the label space ($M_i$ denotes the i-th bit of $M$), where $f_i$ denotes a single feature. Then, a member $l\in L$ is a set of attribute features, which is regarded as a specific kind of label in the following parts. For example, we have 40 attribute features in CelebA~\cite{liu2015faceattributes}. $l$ can be a combination of features such as age, gender, and hair color.
A certain $l$ can determine its label-specific attribute space $A_l$, which is encoded by attribute encoder, and its content space $C_l$, which is encoded by content encoder. An element $y$ denotes a certain specific evaluation of the chosen features in $l$. For example, if $l$ represents hair color and gender in CelebA, then $y$ can be ``black hair male,'' ``blond hair female,'' and so on. 

Let $R_l$ denote label-relevant space which consists of features that label $l$ mentions and $I_l$ denote label-irrelevant space which consists of features that label $l$ does not mention, \textit{i.e.} $I_{l}=\complement_U R_l$, while the label-specific attribute feature space $A_{l}$ is the output of the attribute encoder $E_{a}$, and the content feature space $C_l$ is the output of the content encoder $E_c$.
Then we have four formulas as follows: 

\begin{equation}
    R_l \subset A_l
    \label{formula:RlAl}
\end{equation}
\begin{equation}
    A_l \cup C_l = U
    \label{formula:AlClU}
\end{equation}
\begin{equation}
    R_l \cap I_l = \emptyset
    \label{formula:RlIl}
\end{equation}
\begin{equation}
    R_l \cup I_l = U
    \label{formula:RlIlU}
\end{equation}
where $U$ denotes the space of all features, and the Formulas~\ref{formula:RlIl} and~\ref{formula:RlIlU} holds by our definition. Part ($A$) of Figure~\ref{fig:other_reg} guarantees Formula~\ref{formula:AlClU}. Since the input $y_m$ can be missing in some or all features, to reconstruct $\hat{x}_a$, the attribute features must be provided {\color{black}by $E_a$ and by $E_c$}.
Thus, the union of $A_l$ and $C_l$ must contain all the features. This leads to the Formula~\ref{formula:AlClU}.
An identity reconstruction loss $\mathcal{L}_{rec}$ is shown in Part ($A$).
Using this loss between the input and output images when inputting a matching label can encourage the generator to reconstruct the contents (\textit{e.g.,} background) from the input image.

\begin{equation}
    \mathcal{L}_{rec}=E_x [||G(E_c(x),E_a(x),y_m)-x||_1]
\end{equation}

Figure~\ref{fig:schematic} guarantees Formula~\ref{formula:RlAl}, the label-relevant feature $R_{l}$ is inside the output of the attribute encoder $A_l$, which can be proved by the following 
contradiction:
supposing a feature satisfies $f\in R_l, f\notin A_l$, \textit{i.e.} it is not passed to the final synthesized image $\hat{x}$ through $E_a$. Since the input label $y_m$ can be missing in all features, $f$ must be passed through $E_c$ to reconstruct $\hat{x}$. However, $f$ will be modified by input $y_r$ in the generating process of $\bar{x}$, which will give incorrect attribute features to the generator at the right of Figure~\ref{fig:schematic} to reconstruct $\hat{x}$. This leads to a contradiction. Therefore, Formula~\ref{formula:RlAl} holds. As a result, we can see that the $\mathcal{L}_1$ cycle loss in Figure~\ref{fig:schematic} plays an important role in the label-specific attribute feature disentanglement.
\begin{equation}
\begin{aligned}
    \mathcal{L}_{cyc} = 
    &E_x [||\hat{x}-x||_1],\\
\end{aligned}
\end{equation}
where $\hat{x}=G(E_c(\bar{x}),E_a(x),y_m)$. 
By reconstructing the modified image $\bar{x}$ to the original image $x$ using the attribute embedding from the attribute encoder $E_a$ and the matching label $y_{m}$, we force the output of attribute encoder to contain the label-specific attribute features. The cycle loss also plays an important role in maintaining label-irrelevant contents (\textit{e.g.,} background) from the input image.

Besides, 
the relationship among the different feature spaces mentioned above is shown by Venn diagrams in Figure~\ref{fig:relationship}.
Part ($B$) of Figure~\ref{fig:other_reg} encourages the translation from kind-1 to kind-2, since Part ($B$) requires that the features encoded by $E_a$ correspond to the input noise only. As a result, the intersection of $A_l$ and $C_l$ will decline. 
Finally, when $\mathcal{L}_{latent}$ loss is optimized to $\mathcal{L}_{latent} \rightarrow 0$,
Part ($B$) will lead to the following formula and attain kind-2 relationship,
\begin{equation}
    C_l \subset I_l.
\end{equation}
This achieves label-specific attribute feature disentanglement. The $\mathcal{L}_1$ loss in Part ($B$) can be formulated as follows:
\begin{equation}
    \mathcal{L}_{latent}=E_{x,z}[||E_a(\hat{x}_b)-z||_1]
\end{equation}
where $\hat{x}_b=G(E_c(x),z,y_0)$ and $z\sim \mathcal{N}
(0,1)$ is a random Gaussian noise vector.
By mapping the attribute feature into a prior Gaussian distribution, we enable the manipulation by walking on the attribute feature space, \textit{e.g.,} changing the hair color from one to another smoothly rather than by an immediate change.


\section{Additional Details for Fine-Grained Label Representation}
To better describe the label representation mentioned in the paper, Figure~\ref{fig:sup_label} shows all the possible cases of labels of a face image, which can be viewed as a supplementary of Figure 2 in the paper. Take a brown hair young female in CelebA~\cite{liu2015faceattributes} as an example, you can see that the $y_{GT}$ and $y_0$ both have a unique case, the $y_m$ has 8 cases and the $y_r$ has 36 cases. For each random label $y_{r}$ in the middle table, we can see its corresponding $y_{r} \oplus y_{GT}$ in the right table. 

\section{Additional Experimental Results}
Figure~\ref{fig:sup2} shows additional results of our four different applications generated by DLGAN on CelebA and DukeMTMC.




\ifCLASSOPTIONcaptionsoff
  \newpage
\fi

\end{document}